# Generating Conditional Probabilities for Bayesian Networks: Easing the Knowledge Acquisition Problem

*Balaram Das[1]*
*Command and Control Division, DSTO, Edinburgh, SA 5111, Australia*

**Abstract-** The number of probability distributions required to populate a conditional probability table (CPT) in a Bayesian network, grows exponentially with the number of parent-nodes associated with that table. If the table is to be populated through knowledge elicited from a domain expert then the sheer magnitude of the task forms a considerable cognitive barrier. In this paper we devise an algorithm to populate the CPT while easing the extent of knowledge acquisition. The input to the algorithm consists of a set of weights that quantify the relative strengths of the influences of the parent-nodes on the child-node, and a set of probability distributions the number of which grows only linearly with the number of associated parent-nodes. These are elicited from the domain expert. The set of probabilities are obtained by taking into consideration the heuristics that experts use while arriving at probabilistic estimations. The algorithm is used to populate the CPT by computing appropriate weighted sums of the elicited distributions. We invoke the methods of information geometry to demonstrate how these weighted sums capture the expert's judgemental strategy.

**Key terms:** Bayesian Network, differential geometry, heuristics in judgement, information geometry.

## I. INTRODUCTION

Consider an example where we wish to assess the 'efficiency' of a small business company. Considering the uncertainties involved in such an endeavour, one approach would be to adopt the methods of probabilistic reasoning where we think of 'efficiency' modelled as a random variable $E$, and assign the following five mutually exclusive states to it – very-low ($vl$), low ($l$), average ($a$), high ($h$), and very-high ($vh$). Our assessment would then be in terms of a probability distribution over these states. Now, reasoning about the efficiency in its most general terms would be intractable. An astute analyst would first identify a context or a set of factors that is directly relevant to the kind of assessment desired. Let us say that we can have a reasonable assessment if we consider 'efficiency' as directly dependent upon three factors, these being the levels of – 'personnel morale', 'personnel training', and 'managerial expertise' - in the company. These three factors affecting 'efficiency' can be modelled as random variables $PM$, $PT$, and $ME$ respectively, each having the above mentioned five states. An assessment of these three factors will in turn provide an assessment of 'efficiency'. A rigorous way to do this is through a Bayesian network [1], [2] as shown in Fig. 1.

---

[1] Email: balaram.das@phenomenologist.org

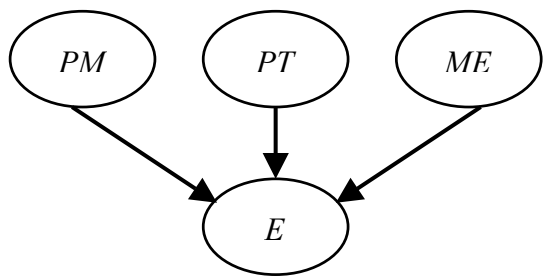

Fig. 1. A Bayesian network to assess the 'efficiency' of a small business company. The child-node *E* representing 'efficiency' is influenced by three parent nodes representing - 'personnel morale' (*PM*), 'personnel training' (*PT*), and 'managerial expertise' (*ME*) - in the company. Each node has 5 states: very-low (*vl*), low (*l*), average (*a*), high (*h*), and very-high (*vh*).

This is a simple Bayesian network with one child-node being influenced by three parent-nodes. In general useful Bayesian networks boast of many more nodes with multiple levels of parent-child dependencies. For example, to assess the state of *PT* one may need to assess a set of other indicators which decide the level of 'personnel training'. These would then form a set of parent-nodes for the node *PT* and so on. It is not our aim here to build a realistic network to assess 'efficiency'. The network in Fig. 1 only serves to illustrate our analysis.

Given our simple example, one assesses the levels of 'personnel morale', 'personnel training', and 'managerial expertise' as probability distributions over the respective random variables. The network accepts these three distributions as input and provides an assessment of 'efficiency' in terms of a probability distribution over *E*. For the network to provide such an output the nature of dependency of the child on its parents needs to be quantified. We do this by attaching a conditional probability table (CPT) to the bunch of arrows linking the child to its parents.

Let us denote by Π a typical *parental configuration* for the network in Fig. 1. More precisely Π is a set consisting of 3 elements, each element representing a state of a different parent such as {*PM* = *a*, *PT* = *l*, *ME* = *l*}. The CPT consists of a collection of probability distributions over the child-node, one for each different parental configuration. These distributions, quantify the parent-child dependency, and have the form:

{$p(E = vl \mid \Pi), p(E = l \mid \Pi), p(E = a \mid \Pi), p(E = h \mid \Pi), p(E = vh \mid \Pi)$}

Here *p(A)* denotes the probability of an event *A* and *p(A|Π)* denotes the conditional probability *p(A| PM = a, PT = l, ME = l)* when Π refers to the parental configuration {*PM* = *a*, *PT* = *l*, *ME* = *l*}.

As there are $5^3$ different parental configurations Π, the CPT will comprise of $5^3$ probability distributions. This large number of distributions demands a considerable amount of intensive effort on the part of an analyst who wishes to generate the CPT. The vexing part is that it is not just large but exponentially large. A CPT quantifying the dependency on *n* parents would demand $5^n$ distributions in order to be functional. It is this exponential growth with the number of parents that constitutes the essential problem.

If we could have a sufficiently adequate database, recording past trends of company efficiency and how it relied on various factors like training and morale etc., then a possible solution would be to automate the process of CPT generation by *learning* the conditional probabilities from the database - a process also referred to as batch learning [3]. However, as Jensen remarks [2, p. 81], "although the statistical theory (for learning) is well understood, no method has so far become the standard". Then again, for many practical problems one rarely finds a relevant database, even an inadequate one at that. In such cases the only available source of probabilistic information are the domain experts.

To elicit required knowledge from the domain expert we have to ask questions of the following type:
*Given the scenario depicted in Fig. 1 and a parental configuration Π, what should be the probability distribution over the states of E?*

We need to ask the question $5^3$ times, once for each different parental configuration Π. We would, say, start with the parental configuration where all the parents are in the state *vl*, figure out the distribution over *E* for this parental configuration, and then repeat this process systematically through all possible parental configurations until we have worked out the distribution over *E* for the configuration where all the parents are in the state *vh*. In this process the distributions that were worked out consecutively would mostly be consistent with each other. But it is well nigh impossible to ensure that distributions that occurred further apart would be mutually consistent. This is essentially because the expert is not a machine and therefore lacks a machine's uncompromising regularity. Just boredom and fatigue during an extended process will ensure that the criteria employed to figure out the distributions are not applied uniformly each time. One way out would be to systematise the method of eliciting probabilistic information from the expert. A number of techniques have been suggested to elicit such information in a reasonable amount of time [4], [5]. However these do not ease the problem of magnitude, the problem of the exponentially large number of probabilities that is required. Domain experts we encounter in model building operate under time constraints and are rarely keen to work through a large list of distributions no matter how little time and effort each distribution requires. As Druzdzel and van der Gaag remark in a fine review [6]: "We feel, therefore, that research efforts aimed at reducing the number of probabilities to be ⋯ of more practical significance."

Some well known methods which reduce the number of probabilities that need to be elicited, are the Noisy-OR model proposed by Pearl [1] and its generalisations by other workers [7]-[9]. These models can compute the distributions needed for the CPT from a set of distributions, elicited from the expert, the magnitude of which grows linearly with the number of parents. However all these models are constrained by the assumption that parents act independently without synergy [7], [8]. This means that parents individually influence the child and that there are negligible cross – interactions between individual parent-to-child influences [1, pp 184-187]. Recently Lemmer and Gossink [10], have advanced the Noisy-OR technique to deal with networks that are not constrained by the above independence requirement. However they constrain their networks to have nodes with binary states only.

For Bayesian networks we are interested in, parents seldom act independently and the nodes often possess more than two states. This paper presents a formalism to tackle such networks. In Section II, we analyse some difficulties associated with knowledge elicitation from a domain expert. This leads us to the concept of compatible parental configurations which we exploit to formulate an algorithm to generate the CPT. The algorithm takes as input (a) a set of weights that quantify the relative strengths of the influences of the parents on the child node and (b) a set of probability distributions over the child node for compatible parental configurations. Thankfully, the latter set grows linearly rather than exponentially with the number of parents. It is difficult to validate the algorithm by comparing its results with similar results obtained from other sources, because such other sources very rarely exist. In Section III we argue that the algorithm would be justified if we can show that the logic behind the algorithm mirrors the judgemental strategy of the domain experts. We employ methods of information geometry to do just this. Finally Section IV concludes the paper.

Software implementing the algorithm currently exists and has been used extensively to construct Bayesian networks for defence related problems [11]-[13].

## II. EASING THE KNOWLEDGE ACQUISITION PROBLEM

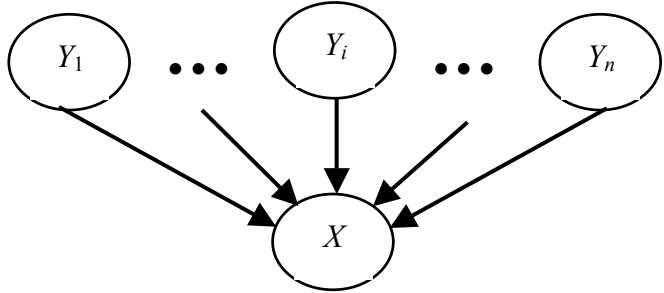

Fig. 2. A Bayesian network with one child-node $X$ with states $\{x^0, x^1, \cdots, x^m\}$ being influenced by $n$ parent-nodes $Y_1, \cdots, Y_n$. Any such parent-node $Y_i$ has $k_i$ states $\{y_i^1, \cdots, y_i^{k_i}\}$ with $k_i \geq 2$.

Let us fix a general network as shown in Fig. 2 to conduct our analysis. The network consists of one child node being influenced by $n \geq 1$ parent nodes. The parent nodes represent $n$ random variables $Y_1, \cdots, Y_n$ and any such variable $Y_i$ has $k_i$ states $\{y_i^1, \cdots, y_i^{k_i}\}$ with $k_i \geq 2$. The child node represents a random variable $X$ with $m+1$ states $\{x^0, x^1, \cdots, x^m\}$ with $m \geq 1$.

A typical *parental configuration* of the network, is a set consisting of $n$ elements, and will take the form:

$\{Y_1 = y_1^{s_1}, \cdots, Y_n = y_n^{s_n}\}$; where $1 \leq s_1 \leq k_1, \cdots, 1 \leq s_n \leq k_n$. (1)

The network will have $k_1 \times \cdots \times k_n$ such parental configurations requiring a CPT with as many probability distributions over the child node *X*. For the parental configuration (1) such a distribution has the form:

$$\{p(x^0 \mid y_1^{s_1}, \cdots, y_n^{s_n}), p(x^1 \mid y_1^{s_1}, \cdots, y_n^{s_n}), \cdots, p(x^m \mid y_1^{s_1}, \cdots, y_n^{s_n})\} \quad (2)$$

Here $p(x^i \mid y_1^{s_1}, \cdots, y_n^{s_n})$ is short for the conditional probability $p(X = x^i \mid Y_1 = y_1^{s_1}, \cdots, Y_n = y_n^{s_n})$; if $\Pi$ denotes the parental configuration (1) then we may also write this conditional probability as $p(x^i \mid \Pi)$.

To elicit the distributions of the type (2) we have to ask the domain expert questions of the following type:

> QA: *Given a parental configuration* $\{Y_1 = y_1^{s_1}, \cdots, Y_n = y_n^{s_n}\}$ *what should be the probability distribution over the states of the child X?*

This question has to be asked $k_1 \times \cdots \times k_n$ times to account for all possible parental configurations. As we have already said, when *n* is large, experts baulk at such an extensive cognitive load. What is more, this is not all that is problematic here. A closer look reveals fundamental difficulties that afflict some of the questions of the type QA. Let us elucidate.

*A. Heuristics in Probability Assessment*

A domain expert's knowledge is gained through personal experience. This knowledge would be as rich as his experiences are and as coloured as his prejudices. The probabilities we extract by questioning such an expert are *subjective* probabilities [3], [14], [15]. For our purposes it is sufficient to accept that this notion of probability is personalistic, it reflects the degree of belief of a particular person at a particular time. However to continue with our deliberations, it is imperative that, we have some understanding as to how people arrive at probabilistic estimations. According to widely accepted research people rarely follow the principles of probability when reasoning about uncertain events, rather they replace the laws of probability by a limited number of *heuristics* [16], [17]. In a classic paper, Kahneman and Tversky describe three frequently used heuristics, (i) *representativeness,* (ii) *availability,* and (iii) *adjustment from an anchor* [18]. Although these heuristics can lead to systematic errors, they are highly economical and usually very effective in making probabilistic judgements.

Let us discuss the *availability* and the related *simulation* heuristics, which are relevant at this point [19]-[21]. Kahneman and Tversky reason as follows. Life-long experience has taught us that frequent events are easier to recall than infrequent ones. The availability heuristics exploits the inverse form of this empirical observation. The more cognitively accessible an event is, the more likely it is perceived to occur. In other words one relates the likelihood of an event to the ease with which this event is recalled or pictured in one's mind. A person is said to employ the availability heuristic whenever he or she estimates the probability of an event by estimating the ease with which the occurrence of the event could be brought to mind. Often one encounters situations where instances of the occurrence of one or other outcome are not stored in the memory ready for recall. Probabilistic judgements for such outcomes are arrived at through a mental operation that resembles the running of a simulation model. One constructs a mental model of the situation and runs a mental simulation to assess the ease with which the simulation could produce different outcomes, given a set of initial conditions and operating parameters. The ease with which the mental simulation reaches a particular outcome is taken as a measure of the propensity of the real system to produce that outcome and hence an assessment of the probability for that outcome, given the initial conditions. This is the simulation heuristic.

Let us apply the above heuristics to our adopted example in Fig. 1. Given that scenario a typical expert would have conceptualised a general set of dependency criteria, albeit subjective, relating personnel wellbeing and competency to organisational efficiency. When asked the question - *given the scenario depicted in Fig.* 1*, and a parental configuration* $\Pi$ *what should be the probability distribution over the states of E*? - the expert may recall instances of companies having parental configurations as in $\Pi$, and how they fared. However, more often than not the expert would carry out a mental simulation. He would start with the initial conditions given by $\Pi$, then using his expert knowledge imagine various paths of evolution of the fortunes of the company. The relative ease with which the simulation produces various states of efficiency can be estimated in a 0 to 1 scale. This in turn would provide the probability distribution over the states of 'efficiency'.

In the more general context questions of the type QA would be answered in a similar manner. However since QA fixes the states of all the parents how would the expert answer the question when the parental states are not consistent with each other? For a better understanding of this point consider the network in Fig. 1 again. Let us fix the parental states as follows: 'personnel morale', is *very-high* 'personnel training' is *very-low*, and 'managerial expertise' is *very-low*. What, we ask the expert, should be the probability distribution over 'efficiency' in this context. Now, one would be hard put to name companies where personnel have very low level of training and managers are highly incompetent but morale is very high nonetheless. Given such a context, the expert will have great difficulty in running mental simulations to imagine the various possible paths of evolution of the fortunes of the company. The specified initial state significantly diverges from normally held belief. According to Kahneman and Tversky this calls for change in one's current mental model before the run of the simulation. The expert's mental model, of the workings of business concerns, would have to be revised first, to make the given context "as unsurprising as possible, and the simulation would employ the parameters of the revised model" [21]. Invariably experts would fail to make such drastic changes to their belief system with equanimity; the simulations they would run would only give rise to more and more confusion and little clarification. To obviate such difficulties we proscribe asking questions of the type QA, which involve incompatible parental states. In fact we adopt a very different strategy, one that directly expunges such eventualities.

### B. Compatible Parental Configurations

To restate our problem, some parental configurations would seem incompatible to the expert when his mental model does not support them. How would the expert deal with such parental configurations? Interestingly, such situations do not arise for those networks that conform to the Noisy-OR model because of the independency assumption used therein. It is therefore instructive to look at an example, and Pearl gives a good one. He considers a scenario where a burglar alarm can be triggered either by an attempted burglary or by an earthquake [1, pp 184-187]. This scenario can be modelled as a Bayesian network where the burglar alarm is the child-node, and this is being influenced causally by two parent-nodes representing burglary and earthquake. As Pearl correctly argues, reasoning about burglaries and reasoning about earthquakes belong to disparate frames of knowledge - frames of knowledge that are truly foreign to one another. Hence the individual parental influences, in this case, are necessarily independent of each other - a fact that makes the Noisy-OR and its generalised versions work. The same premises would also ensure that parental configurations never appear to be incompatible to an expert, as he would be reasoning about individual parents in disparate frames of knowledge. This is not true for the networks we have under present consideration. In other words for networks that interests us, there necessarily exists a coherent frame of knowledge which confines within itself reasoning about each and every parent and their influences on the child. Given our example in Fig. 1, say, a typical expert would have studied the rise and fall of various companies. He would have noted how their fortunes were decided by the capabilities of their personnel and how these capabilities naturally interact with each other. In other words any competent expert would duly acquire a frame of knowledge within which his reasoning about the parents and their influences can be conducted with little difficulty. This motivates our two assumptions below.

*Definition*: Consider the network in Fig. 2, and suppose that the parent $Y_i$ is in the state $y_i^{s_i}$. We say that the state $Y_j = y_j^{s_j}$, for the parent $Y_j$, is *compatible* with the state $Y_i = y_i^{s_i}$, if according to the expert's mental model the state $Y_j = y_j^{s_j}$ is most likely to coexist with the state $Y_i = y_i^{s_i}$. Let $\{Comp(Y_i = y_i^{s_i})\}$ denote the *compatible parental configuration* where $Y_i$ is in the state $y_i^{s_i}$ and the rest of the parents are in states compatible with $Y_i = y_i^{s_i}$.

We now make our first assumption:

*Assumption A*: Given the network in Fig. 2, the expert can acquire a coherent frame of knowledge within which it is possible to decide upon a compatible parental configuration $\{Comp(Y_i = y_i^{s_i})\}$ for any parent $Y_i$ and any state $y_i^{s_i}$ of this parent.

Clearly different experts will conceptualise $\{Comp(Y_i = y_i^{s_i})\}$ in different ways. Instead of asking questions of the type QA, we now ask questions of the following type:

> QB: *Given the parental configuration $\{Comp(Y_i = y_i^{s_i})\}$ what should be the probability distribution over the states of the child X?*

In other words we seek distributions of the type:

$$\{p(x^0 \mid \{Comp(Y_i = y_i^{s_i})\}), p(x^1 \mid \{Comp(Y_i = y_i^{s_i})\}), \cdots, p(x^m \mid \{Comp(Y_i = y_i^{s_i})\})\}$$
where $1 \leq i \leq n$ and $1 \leq s_i \leq k_i$. (3)

*Remark*: Let us stress that $\{Comp(Y_i = y_i^{s_i})\}$ is a parental configuration in the mental model of the expert where he has chosen to focus on the state $y_i^{s_i}$ of the parent $Y_i$ while the rest of the parents are perceived in his judgement to be in states compatible with this state of $Y_i$. We are not asserting that parental configurations are only to be found as compatible parental configurations in reality. Nor are we asserting that questions of the type QA with manifestly incompatible parental configuration are meaningless questions. The concern is rather with the fact that when the expert believes a certain parental configuration to be incompatible, QA would lead to confusion. Why would a divergence like this exist between the expert's reasoning and the reasoning required to populate the CPT? The explanation is quite simple. The expert's reasoning is guided not only by the nature of the network but also by the ambient knowledge that impinges on the network. The CPT, on the other hand, is separated from the ambient knowledge by the requirements of *d-separation* [1], [2], which Bayesian networks must satisfy. Our mode of knowledge elicitation must conform to the expert's way of thinking, as he is our primary source of knowledge. We ensure this by asking questions of the type QB, which involve, only compatible parental configurations as perceived by the expert. How to be in conformity with the expert and still be able to generate the CPT is what we are going to discuss next.

Our Bayesian network will have $k_1 + \cdots + k_n$ compatible parental configurations of the type $\{Comp(Y_i = y_i^{s_i})\}$ requiring us to ask the question QB that many times to obtain all possible distributions of the type (3). Note that the number of distributions of the type (3) grows linearly rather than exponentially with *n*. For example, with the network shown in Fig. 1 there are 5×3 distributions of the type (3) as against $5^3$ distributions of the type (2).

If the states of the parents have a one-to-one compatibility correspondence, which is also an equivalence relation, then the number of questions of the type QB can be reduced drastically. Let us explain. Consider the case where the parents have the same number of states i.e., $k_1 = k_2 = \cdots = k_n = k$, and suppose that:

$Y_i = y_i^t$ is compatible with $Y_j = y_j^t$
for $1 \le t \le k$ and for any *i* and *j* in $\{1, \cdots, n\}$, but (4)
$Y_i = y_i^t$ is not compatible with $Y_j = y_j^s$
whenever $t \ne s$

Then using the symbol '≡' to relate two identical sets, we have

$\{Comp(Y_1 = y_1^t)\} \equiv \{Comp(Y_2 = y_2^t)\} \equiv \cdots \equiv \{Comp(Y_n = y_n^t)\}$
$\equiv \{Y_1 = y_1^t, Y_2 = y_2^t, \cdots, Y_n = y_n^t\}$ for $1 \le t \le k$.

Hence there are only *k* different compatible parental configurations corresponding to the *k* states of any parent. In other words we have to ask the question QB *k* times only, or the number of questions, in this case, are fixed and do not grow with the number of parents. One should therefore endeavour to formulate the problem such that the compatibility correspondences as in (4) are achieved. It may not always be possible to achieve one-to-one correspondence among all the parents. Nonetheless, making it happen among a subset of parents will help reduce the number of questions to be asked. We illustrate this with an example below.

*1) Example*: Consider the network depicted in Fig. 1. To assign the 5×3 probability distributions, of the type (3), over the child node *E*, for compatible parental configurations, let us say the expert adopts the following way of thinking. He starts with the parent *PM*, for which he subjectively interprets the compatible parental configurations as follows:

$\{Comp(PM = s)\} \equiv \{PM = s, PT = s, ME = s\}$, for $s = vl, l, a, h, vh$.

If one treats compatibility as an equivalence relation then, to be consistent, the subjective interpretation for the compatible parental configurations related to other parents should be similar i.e.,

$$\{Comp(PM = s)\} \equiv \{Comp(PT = s)\} \equiv \{Comp(ME = s)\}$$
$$\equiv \{PM = s, PT = s, ME = s\}, \quad \text{for } s = vl, l, a, h, vh \quad (5)$$

Hence for the probability distribution over the child node *E* one will have:

$$p(E = e \mid \{Comp(PM = s)\})$$
$$= p(E = e \mid \{Comp(PT = s)\})$$
$$= p(E = e \mid \{Comp(ME = s)\}) \quad \text{for } e, s = vl, l, a, h, vh \quad (6)$$

In other words if the expert provides the 5 probability distributions over the node *E* corresponding to the 5 parental configurations $\{Comp(PM = s)\}$, $s = vl, l, a, h, vh$, then all the 5×3 distributions for compatible parental configurations are obtained.

Achieving an agreeable state of affairs as the one just discussed would not always be possible. There are many factors which can deny the required one-to-one correspondence. An obvious example is one where parents have a different number of states. To see what can go wrong, consider Fig. 1 again, now with the proviso that the node *PT* has only three states $\{l, a, h\}$. For such a situation an expert may assign compatibility as follows:

$\{Comp(PM = vl)\} \equiv \{Comp(ME = vl)\} \equiv \{PM = vl, PT = l, ME = vl\}$;
$\{Comp(PM = l)\} \equiv \{Comp(ME = l)\} \equiv \{PM = l, PT = l, ME = l\}$;
$\{Comp(PM = a)\} \equiv \{Comp(ME = a)\} \equiv \{PM = a, PT = a, ME = a\}$;
$\{Comp(PM = h)\} \equiv \{Comp(ME = h)\} \equiv \{PM = h, PT = h, ME = h\}$;
$\{Comp(PM = vh)\} \equiv \{Comp(ME = vh)\} \equiv \{PM = vh, PT = h, ME = vh\}$.

The state *PT = l* occurs in both the first two parental configurations above, thus we have lost the one-to-one correspondence. If, for the sake of argument, the expert perceives

$\{Comp(PT = l)\} \equiv \{PM = vl, PT = l, ME = l\}$,

then we have to add this to the 5 parental configurations listed above and a similar addition may be required for the state *PT = h*. Thus we would require more than 5 probability distributions over the node *E*. Possibly 7, but still less than 15 because we have a one-to-one correspondence between the states of *PM* and *ME*.

*C. The Relative Weights*

The CPT nevertheless requires probability distributions for all possible parental configurations compatible or not. Is there a way to generate the rest of the distributions from the compatible ones? This leads us to the concept of relative weights.

If the expert can acquire a coherent frame of knowledge, within which he can reason about all the parental influences, then this ability in turn would give the ability to compare these influences - i.e. an ability to figure out which parents are more influential and which are less so. Experience in populating CPTs [11]-[13], shows that experts happily exploit this ability when they can. They find it instinctive to assign relative weights to the parents to quantify the relative strengths of their influences on the child node. Weights, it seems, assist the process of mental simulations as discussed before. We therefore make our second assumption.

*Assumption B*: Given the network in Fig. 2, the expert can acquire a coherent frame of knowledge to reason about the parental influences such that it is possible to assign relative weights $w_1,\cdots,w_n$, to the parents $Y_1,\cdots,Y_n$ respectively, to quantify the relative strengths of their influences on the child node *X*. Of course the weights have to be positive and for later convenience we stipulate that they be assigned in a normalised form i.e., $0 \leq w_i \leq 1$, for $i = 1,\cdots,n$, and $w_1 + \cdots + w_n = 1$.

*D. The weighted sum algorithm*

If all the information that the expert is willing to give us are:

(a) the relative weights $w_1,\cdots,w_n$ and
(b) the $k_1 + \cdots + k_n$ probability distributions over *X*, (7)
    of the type (3), for compatible parental configurations,

then can we generate the CPT for the network in Fig. 2? The answer is 'yes' and we put forward the following algorithm.

The relative weights naturally suggest a weighted sum procedure. Given the information contained in (7), we propose the following weighted sum algorithm to estimate the $k_1 \times \cdots \times k_n$ distributions of the type (2).

$$p(x^l \mid y_1^{s_1}, y_2^{s_2},\cdots,y_n^{s_n}) = \sum_{j=1}^{n} w_j p(x^l \mid \{Comp(Y_j = y_j^{s_j})\}) \qquad (8)$$

here $l = 0,1,\cdots,m$ and $s_j = 1,2,\cdots,k_j$.

*1) Example:* Before we examine the validity of the weighted sum algorithm let us check if it is intuitively satisfying as per the network in Fig. 1. To start with, the expert assigns the parent nodes *PM, PT,* and *ME* with the relative weights $w_1, w_2$ and $w_3$ respectively. Next he specifies the 5×3 probability distributions, of the type (3), over the child node *E* for compatible parental configurations. If he adopts the way of thinking described in Section II-*B-1* then all he needs to do is provide the 5 probability distributions over the node *E* corresponding to the 5 parental configurations {*Comp(PM = s)*}, *s = vl, l, a, h, vh*. Let us say the expert estimates these distributions as given in Table 1. The essence of this estimation is that the parental configuration {*Comp(PM = s)*} gives rise to a unimodal probability distribution over *E* with the mode at *E = s*.

TABLE 1
Distribution over *E* for compatible parental configurations {*Comp*(*PM* = *s*)}

| Probability distribution over *E* | *s* = *vl* | *s* = *l* | *s* = *a* | *s* = *h* | *s* = *vh* |
|---|---|---|---|---|---|
| *p*(*E* = *vl* | {*Comp*(*PM* = *s*)}) | 0.8 | 0.08 | 0.02 | 0.01 | 0.005 |
| *p*(*E* = *l* | {*Comp*(*PM* = *s*)}) | 0.15 | 0.8 | 0.08 | 0.03 | 0.015 |
| *p*(*E* = *a* | {*Comp*(*PM* = *s*)}) | 0.03 | 0.08 | 0.8 | 0.08 | 0.03 |
| *p*(*E* = *h* | {*Comp*(*PM* = *s*)}) | 0.015 | 0.03 | 0.08 | 0.8 | 0.15 |
| *p*(*E* = *vh* | {*Comp*(*PM* = *s*)}) | 0.005 | 0.01 | 0.02 | 0.08 | 0.8 |

From the 5 distributions in Table 1 all the 5×3 distributions of the type (3) can be obtained using relations (6). Taking these distributions and the weights as input, the weighted sum algorithm (8) would be able to calculate all the $5^3$ distributions required to populate the CPT. Let us consider two instances of this calculation to see if the algorithm calculates intuitively satisfying distributions.

*Compatible Parental Configuration*: Consider first the simple case where we want to find the distribution over *E* when *PM*, *PT*, and *ME* are all in the state *vh* i.e.:

$$\{p(E = e \mid PM = vh, PT = vh, ME = vh), e = vl, l, a, h, vh \} \quad (9)$$

According to (8) we need to carry out the following computations

$$\begin{aligned} &p(E = e \mid PM = vh, PT = vh, ME = vh) \\ &= w_1 p(E = e \mid \{Comp(PM = vh)\}) + w_2 p(E = e \mid \{Comp(PT = vh)\}) \\ &\quad + w_3 p(E = e \mid \{Comp(ME = vh)\}), e = vl, l, a, h, vh. \end{aligned} \quad (10)$$

Taking the relation (6) and the fact that $w_1 + w_2 + w_3 = 1$ into account, (10) gives:

$$p(E = e \mid PM = vh, PT = vh, ME = vh)$$
$$= p(E = e \mid \{Comp(PM = vh)\}), e = vl, l, a, h, vh.$$

But this is what we expected because of the way of thinking adopted by the expert regarding compatible parental states as recorded in relation (5). More explicitly the distribution (9) corresponds to the last column of Table 1.

*Incompatible Parental Configuration*: Let us next consider the parental configuration where 'personnel morale' is in the state *very-high* while 'personnel training', and 'managerial expertise' are each in the state *very-low*. We seek the distribution

$$p(E = e \mid PM = vh, PT = vl, ME = vl), e = vl, l, a, h, vh \quad (11)$$

According to (8) we need to carry out the following computations

$$\begin{aligned} &p(E = e \mid PM = vh, PT = vl, ME = vl) \\ &= w_1 p(E = e \mid \{Comp(PM = vh)\}) + w_2 p(E = e \mid \{Comp(PT = vl)\}) \\ &\quad + w_3 p(E = e \mid \{Comp(ME = vl)\}), e = vl, l, a, h, vh. \end{aligned} \quad (12)$$

We have discussed this parental configuration in Section II-*A* where we remarked that it promotes confusion in probabilistic assessments, as the configuration diverges from the expert's normally held belief. This normally held belief is reflected in the compatible parental configurations set down in (5), according to which *PM* = *vh* is incompatible with *PT* = *vl* and *ME* = *vl*. The expert's belief is also reflected in the unimodal nature of the distributions in Table 1, which says that {*Comp*(*PM* = *vh*)} has an influence on *E* which contradicts the influences of the parental configurations {*Comp*(*PT* = *vl*)} and {*Comp*(*ME* = *vl*)}. Suppose the expert assigns the weights as - $w_1 = 0.5, w_2 = 0.25, w_3 = 0.25$. In other words he considers *PM* to be as influential as *PT* and *ME* combined. This further sharpens the contradiction. What would be the probability distribution (11) as computed by the algorithm. This is easily found by substituting the weights and the respective distributions from Table 1 into (12). Column 2 of Table 2 displays this distribution. Note that the distribution is bimodal with modes at *E* = *vl* and at *E* = *vh* portraying a confusing state of affairs as would be expected. Let us try another set of weights such as $w_1 = 0.1, w_2 = 0.45, w_3 = 0.45$. This suggests that the influence of *PM* on *E* is small compared to the combined influences of *PT* and *ME*. This weakens the contradiction as we can now overlook the influence of *PM* on *E*. The corresponding distribution in column 3 of Table 2 reflects this less confusing state of affairs.

TABLE 2
Distribution over *E* for the parental configuration {*PM* = *vh*, *PT* = *vl*, *ME* = *vl*}

|  | $w_1 = 0.5, w_2 = 0.25, w_3 = 0.25$ | $w_1 = 0.1, w_2 = 0.45, w_3 = 0.45$ |
|---|---|---|
| *p*(*E* = *vl* \| *PM* = *vh*, *PT* = *vl*, *ME* = *vl*) | 0.4025 | 0.7205 |
| *p*(*E* = *l* \| *PM* = *vh*, *PT* = *vl*, *ME* = *vl*) | 0.0825 | 0.1365 |
| *p*(*E* = *a* \| *PM* = *vh*, *PT* = *vl*, *ME* = *vl*) | 0.03 | 0.03 |
| *p*(*E* = *h* \| *PM* = *vh*, *PT* = *vl*, *ME* = *vl*) | 0.0825 | 0.0285 |
| *p*(*E* = *vh* \| *PM* = *vh*, *PT* = *vl*, *ME* = *vl*) | 0.4025 | 0.0845 |

Apart from being intuitively satisfying the weighted sum should give us a valid algorithm. This question of validity is rather a tricky one. We are not trying to validate the CPT produced by the weighted sum algorithm against a CPT that can be produced through other means e.g., one that can be produced through batch learning. This is for the obvious reason that such other means of producing a CPT seldom exist, due to the reasons already discussed. Rather the right question to ask is this.

> Suppose all the information that the expert is willing to give us are those contained in (7). Then, does the weighted sum procedure (8) provide an algorithm that is commensurate with the expert's subjective judgement?

In the next section we will show that the answer to this question is in the affirmative.

# III. JUSTIFICATION OF THE WEIGHTED SUM ALGORITHM

## A. Heuristics Governing the Algorithm

Our algorithm takes in the information contained in (7) as input and generates the CPT. Suppose we ask the expert to do the same. In other words after eliciting the information contained (7) we ask the expert to build upon this information and generate the CPT on his own. How would he set about this task? Our motive in asking this rather rhetorical question is the following. If we can show, in a mathematically valid way, that our algorithm mirrors the judgemental strategy adopted by the expert in accomplishing the above task, then we will consider our algorithm to have been justified.

To understand how the expert would proceed starting from (7), let us consider an example. To make it simple, let us say that our network in Fig. 2 has only two parent-nodes representing random variables $Y_1$ and $Y_2$. We ask the expert for the relative weights $w_1$ and $w_2$ and the set of $k_1 + k_2$ probability distributions over $X$ for compatible parental configurations as in (13) and (14) below:

$$\{p(x^0 | \{Comp(Y_1 = y_1^{s_1})\}), p(x^1 | \{Comp(Y_1 = y_1^{s_1})\}), \cdots, p(x^m | \{Comp(Y_1 = y_1^{s_1})\})\}$$
$$1 \leq s_1 \leq k_1 , \tag{13}$$

and

$$\{p(x^0 | \{Comp(Y_2 = y_2^{s_2})\}), p(x^1 | \{Comp(Y_2 = y_2^{s_2})\}), \cdots, p(x^m | \{Comp(Y_2 = y_2^{s_2})\})\}$$
$$1 \leq s_2 \leq k_2 . \tag{14}$$

With this information as a set of building blocks the expert now proceeds to estimate all probability distributions of the type

$$\{p(x^0 | y_1^{s_1}, y_2^{s_2}), p(x^1 | y_1^{s_1}, y_2^{s_2}), \cdots, p(x^m | y_1^{s_1}, y_2^{s_2})\}$$
$$1 \leq s_1 \leq k_1 \text{ and } 1 \leq s_2 \leq k_2 . \tag{15}$$

Recall that $\{Comp(Y_1 = y_1^{s_1})\}$ is a parental configuration in the mental model of the expert where he has chosen to focus on the state $y_1^{s_1}$ of the parent $Y_1$ while $Y_2$ is perceived in his judgement to be in a state compatible with this state of $Y_1$, and (13) gives the distribution over $X$ for all such compatible parental configurations. A similar interpretation goes for (14). Hence for any particular $s_1$ and $s_2$, the expert would use (13) and (14) as reference points to come up with an assessment of (15).

In fact such a judgemental strategy would be an instance of the heuristic *adjustment and anchoring*. According to Kahneman and Tversky [18], "In many situations, people make estimates by starting from an initial value that is adjusted to yield a final answer. The initial value, or starting point, may be suggested by the formulation of the problem, or it may be the result of a partial computation". In our case (13) and (14), for some $s_1$ and $s_2$, are the two initial distributions. Using these as anchors the expert would adjust or interpolate (15), for the same $s_1$ and $s_2$, to lie somewhere in between these two distributions for compatible parental states.

Hence, the intuitive notion of interpolating a probability distribution in between two given distributions, or anchors, must first be rigorously spelt out. For this purpose, let us consider the set $\mathcal{P}$ of all probability distributions over the random variable $X$. Given two distributions $p_1$ and $p_2$ in $\mathcal{P}$, the intuitive notion that another distribution $p$ is interpolated somewhere in between $p_1$ and $p_2$, would mean that $p$ lies on the straight line joining $p_1$ and $p_2$; this of course if $\mathcal{P}$ happened to be a flat space. For $\mathcal{P}$ with more complicated geometry this would mean that $p$ lies on the geodesic joining $p_1$ and $p_2$, for a geodesic encapsulates the notion of straightness. To generalise this concept, suppose we are given $n$ distributions $p_1,\cdots,p_n$ in $\mathcal{P}$. How would we represent the intuitive notion of a probability distribution $p$ being interpolated somewhere in between these given $n$ distributions? Let us assume that $\mathcal{P}$ is endowed with a well behaved geometry such that there exists a unique geodesic connecting any two points in $\mathcal{P}$. Now recall that a subset $C \subseteq \mathcal{P}$ is said to be convex if any two arbitrary points in $C$ can be joined by a geodesic segment contained in $C$ [22]. Hence, the generalisation of the concept of a geodesic joining $p_1$ and $p_2$ is the concept of a convex hull spanned by the $n$ points $p_1,\cdots,p_n$ in $\mathcal{P}$, therefore any interpolated distribution $p$ would lie in such a convex hull.

Let us now revert back to our original network with $n$ parents as shown in Fig. 2. Given any $s_1,\cdots,s_n$, to estimate the probability distribution

$$\left\{p\left(x^0 \mid y_1^{s_1}, y_2^{s_2}, \cdots, y_n^{s_n}\right), p\left(x^1 \mid y_1^{s_1}, y_2^{s_2}, \cdots, y_n^{s_n}\right), \cdots, p\left(x^m \mid y_1^{s_1}, y_2^{s_2}, \cdots, y_n^{s_n}\right)\right\}, \qquad (16)$$

the expert would treat the $n$ elicited distributions

$$\left\{p\left(x^0 \mid \{Comp(Y_j = y_j^{s_j})\}\right), p\left(x^1 \mid \{Comp(Y_j = y_j^{s_j})\}\right), \cdots, p\left(x^m \mid \{Comp(Y_j = y_j^{s_j})\}\right)\right\}$$
$$j = 1, \cdots, n, \qquad (17)$$

for compatible parental configurations, as $n$ anchor points and interpolate (16) to lie somewhere in between these points. Hence if our algorithm were to capture the expert's judgmental strategy it should compute the distribution (16) to lie in the convex hull spanned by the $n$ distributions in (17). Where exactly it should lie in the convex hull would be determined by the $n$ weights $w_1, w_2, \cdots, w_n$.

*B. A Few Concepts from Information Geometry*

Convex sets and geodesics are geometric concepts, what we need therefore is a geometric appreciation of a set of probability distributions. The fact that such an appreciation can be obtained, or more specifically the fact that the methods of differential geometry can be fruitfully applied to gain insight into statistical structures was first realised in 1945 by Rao [23], who pointed out that the Fisher information determines a Riemannian metric on the set of probability distributions. Since then statisticians have enriched this insight [24]-[26] so much so that we now have a sophisticated way of viewing statistical structures in terms of geometry. Amari has coined the phrase *information geometry* to denote this happy confluence of ideas from geometry and statistics. In this section we compile a few information geometric tools necessary for our analysis. In doing so we adhere closely to Amari's treatment [27] of the subject. The reader may also wish to consult [28] for additional insight. The treatment of differential geometric concepts listed below is more in the nature of establishing notations than providing explanations. Adequate explanations can be readily perused in standard texts such as [29], [30].

Consider a set $\mathcal{X}$ having a finite number of elements. By probability distributions over $\mathcal{X}$ we mean functions such as $p : \mathcal{X} \to \mathbb{R}$ such that $p(x) \geq 0$ ($\forall x \in \mathcal{X}$) and $\sum_{x \in \mathcal{X}} p(x) = 1$. Let $S$ be a set of these probability distributions chosen such that each element of $S$ is parameterised by $d$ real parameters; or we require the existence of a mapping $\phi : S \to \mathbb{R}^d$ which is one-to-one. Let us assume $\phi(S)$ to be an open subset in $\mathbb{R}^d$ and let $r^i : \mathbb{R}^d \to \mathbb{R}$, $i = 1, \cdots, d$, denote the $i$th coordinate function on $\mathbb{R}^d$, i.e. it acts on points in $\mathbb{R}^d$ to give their $i$th coordinate. If now we denote $\theta^i = r^i \circ \phi$, then we can view $\phi$ as a global coordinate map on $S$ with coordinate functions $\{\theta^1, \cdots, \theta^d\}$. To convey this parameterisation explicitly we write the probability distributions in $S$ in the form $p(x; \theta^1, \cdots, \theta^d)$ or more compactly as $p(x; \theta)$ where we have substituted $\theta$ for $\{\theta^1, \cdots, \theta^d\}$. To carry out unhindered differentiations with respect to the parameters we assume that, for any $x \in \mathcal{X}$, $p(x; \theta^1, \cdots, \theta^d)$ as a function of $\{\theta^1, \cdots, \theta^d\}$ is $C^\infty$ in its domain of definition, which is $\phi(S)$. One can of course achieve the above parameterisation through various other mappings $\psi : S \to \mathbb{R}^d$. We consider all such mappings for which $\psi \circ \phi^{-1}$ and $\phi \circ \psi^{-1}$ are $C^\infty$ - these will constitute a maximal $C^\infty$ atlas on $S$ that includes $\phi$ thus making $S$ a $C^\infty$ $d$-dimensional manifold.

To formulate the concepts of geodesics and convex sets, one requires the notion of parallel transport to be defined on $S$. This is done by defining an affine connection $\nabla$. In general $\nabla$ is an operator that maps any pair of $C^\infty$ vector fields $U$ and $V$ defined on $S$ to a $C^\infty$ vector field $\nabla_U V$ on $S$ while satisfying the properties of linearity and the Leibnitz rule [29], [30]. For later convenience we seek to explicate this concept with respect to the coordinate basis $\{\theta^1, \cdots, \theta^d\}$ defined above. Given this coordinate basis we can define $d$ vector fields $\partial_i$ on $S$, in the usual manner i.e. for any point $q \in S$, $\partial_i : q \to \left(\dfrac{\partial}{\partial \theta^i}\right)_q \in T_q(S)$ where $T_q(S)$ is the tangent space at $q$. An affine connection $\nabla$ on $S$ is completely specified by specifying the $d^3$ connection coefficients $\Gamma_{ij}^k : S \to \mathbb{R}$ where $i,j,k = 1, \cdots, d$, and $\Gamma_{ij}^k$ are $C^\infty$ functions. The connection $\nabla$ is then specified by the rule $\nabla_{\partial_i} \partial_j = \Gamma_{ij}^k \partial_k$; if in addition we have a metric $g = g_{ij} d\theta^i \otimes d\theta^j$, $i,j = 1, \cdots, d$, defined on $S$, we can use the metric to lower the upper index of the connection coefficients to give us $\Gamma_{ij,k} = \Gamma_{ij}^h g_{hk}$. Note that we have invoked the summation convention where appropriate, i.e., indices appearing as subscripts and superscripts are summed over from 1 to $d$.

*1) Amari's α-connections:* Clearly the connection coefficients and hence a connection ∇ can be assigned in an infinite variety of ways. Each such assignment assigns a geometry to $\mathcal{S}$ that is peculiar to that connection and each such geometry has its characteristic notions of geodesics and convex sets. We cannot have a definite notion of geodesics and convex sets unless we choose a definite connection to structure our manifold $\mathcal{S}$. The question therefore is - are there any connections whose distinguishing properties make them undeniably suitable for our analysis? The answer is in the affirmative and the connections are a family of connections known as α-connections as defined by Amari. Amari and Nagaoka [27] write as follows - "a statistical model $\mathcal{S}$, in addition to its structure as a manifold, has the property that each point denotes a probability distribution. Taking this property into consideration, we find that there are natural structural conditions which are uniquely met by (the Fisher metric and) the α-connections". We sketch here, for quick reference, their arguments for preferring the α-connections; of course readers seeking more detailed reassurance should peruse them in the original. The reasoning runs as follows. In general information is lost when one summarises statistical data. However if the summarization takes place through sufficient statistic the information loss is nil. Therefore the concept of sufficient statistic plays an important role in statistical theories. The α-connections are uniquely characterised by the fact that they remain invariant when we pass from one statistical model to another through sufficient statistic.

Let us proceed to define the α-connections. These are denoted by $\nabla^{(\alpha)}$, and defined through their connection coefficients. With respect to the coordinate basis $\{\theta^1, \cdots, \theta^d\}$ these $d^3$ coefficients are given as:

$$\Gamma_{ij,k}^{(\alpha)}(\theta) = \sum_{x \in \mathcal{X}} \left[ \left( \begin{array}{c} \partial_i \partial_j \log p(x;\theta) + \\ \frac{1-\alpha}{2}(\partial_i \log p(x;\theta))(\partial_j \log p(x;\theta)) \end{array} \right) (\partial_k \log p(x;\theta)) p(x;\theta) \right]; \qquad (18)$$

here $\alpha$ is an arbitrary real number, and $i,j,k = 1, \cdots, d$.

The lowering of the index of the connection coefficients is achieved through the Fisher metric whose coefficients are given as follows:

$$g_{ij}(\theta) = \sum_{x \in \mathcal{X}} \left( (\partial_i \log p(x;\theta))(\partial_j \log p(x;\theta)) p(x;\theta) \right), i,j, = 1, \cdots, d$$

*C The Weighted Sum Algorithm; A Geometric Point of View*

The weighted sum algorithm is concerned with probability distributions over the random variable $X$ with $m+1$ states $\{x^0, x^1, x^2, \cdots, x^m\}$. Hence we take $\mathcal{X} = \{x^0, x^1, x^2, \cdots, x^m\}$. Consider the set $\mathcal{P}$ of all probability distributions over $\mathcal{X}$. We define the global coordinate map $\phi: \mathcal{P} \to \mathbb{R}^m$ in the following form. Given a distribution $p \in \mathcal{P}$, we assign coordinates $\{\theta^1, \cdots, \theta^m\}$ to $p$ as follows:

$\theta^i(p) = r^i \circ \phi(p) = p(x^i)$; where $r^i : \mathbb{R}^m \to \mathbb{R}$, denotes the *i*th coordinate function on $\mathbb{R}^m$, $i = 1, \cdots, m$.

In other words $\phi(\mathcal{P})$ is an open set in $\mathbb{R}^m$ given by

$$\phi(\mathcal{P}) = \left\{ \{\theta^1, \cdots, \theta^m\} \in \mathbb{R}^m \mid \theta^i > 0 \, (\forall i), \text{ and } \sum_{i=1}^{m} \theta^i < 1 \right\}$$

We can convey this coordinate system explicitly by writing the probability distributions as:

$$p(x^l; \theta) = \begin{cases} \theta^l, & (1 \leq l \leq m) \\ 1 - \sum_{i=1}^{m} \theta^i, & (l = 0) \end{cases} \tag{19}$$

One can now provide $\mathcal{P}$ with an $C^\infty$ atlas that includes $\phi$ thus making it an $C^\infty$ *m*-dimensional manifold. This parameterisation $\phi$ is also called a *mixture parameterisation* [27].

From (18) the $\alpha$-connection coefficients in the coordinate system $\{\theta^1, \cdots, \theta^m\}$, can be calculated to give

$$\Gamma_{ij,k}^{(\alpha)}(\theta) = -\sum_{h=1}^{m} \frac{1+\alpha}{2(\theta^h)^2} \delta_i^h \delta_j^h \delta_k^h ; \tag{20}$$

where the traditional Kronecker delta is defined as $\delta_j^i = \begin{cases} 1 \text{ when } i = j \\ 0 \text{ when } i \neq j \end{cases}$.

For each different value of the parameter $\alpha$, we get a different set of connection coefficients, each set imparting a characteristic geometry to $\mathcal{P}$. All we know is that the appropriate connections for statistical manifolds are $\alpha$-connections. But which value of $\alpha$ should characterise the manifold under consideration is not known a-priori. Recall that the information that the expert is willing to give us are all contained in (7). These information are not sufficient to provide us with the specific nature of the connection coefficients on of the manifold $\mathcal{P}$. The prudent course in such circumstances is to opt for the simplest possible geometry that an $\alpha$-connection can impart to $\mathcal{P}$. This is easily done; for $\alpha = -1$, all the connection coefficients in (20) vanish imparting $\mathcal{P}$ with the simplest geometry, that of a flat manifold. We write $\nabla^{(-1)}$ to denote this flat connection.

*1) Flatness as a Mental Model*: Our appeal to the simplest geometry resulted in assigning $\mathcal{P}$ with the flat connection $\nabla^{(-1)}$. Yet as far as the human way of thinking is concerned the flatness of $\mathcal{P}$ is even more natural than that of choosing the simplest connection. Let us portray $\mathcal{P}$ as follows. Let $\mathbb{R}^{\mathcal{X}}$ be the set of all real valued functions defined on $\mathcal{X}$, and consider the subset $F = \left\{ f \in \mathbb{R}^{\mathcal{X}} \mid \sum_{x \in \mathcal{X}} f(x) = 1 \right\} \subseteq \mathbb{R}^{\mathcal{X}}$. It can be shown that $F$ is an affine subspace of $\mathbb{R}^{\mathcal{X}}$ [27, p 40]. As an affine space, $F$ has a natural connection associated with it, which is a flat connection [28, pp 109-111].

Hence the natural geometry on $F$ is a flat geometry. Now the natural embedding of $\mathcal{P}$ into $\mathbb{R}^X$ maps $\mathcal{P}$ into an open subset of the affine space $F$ – in fact it embeds it onto $\{f \in F \mid f(x) > 0 \ \forall x \in \mathcal{X}\}$. Hence as Amari points out, the $\nabla^{(-1)}$ connection on $\mathcal{P}$ is nothing but the natural connection induced from the affine structure of $F$. When humans think of a probability distribution $p$ over the set $\mathcal{X}$ they think of it in terms of a set of $(m + 1)$ real numbers $p(x^l)$, $l = 0, 1, \cdots, m$, which are the probability value associated with $x^l \in \mathcal{X}$. In other words we think of $\mathcal{P}$ as it would appear if it were naturally embedded in $\mathbb{R}^X$. Hence the mental picture of $\mathcal{P}$ we have is naturally associated with the flat connection $\nabla^{(-1)}$, and the linear structure of the weighted sum algorithm is a direct consequence of this mental picture.

*2) Weighted Sums Generate Convex Sets*: Let $\{p_1, \cdots, p_n\}$, be $n$ probability distributions in $\mathcal{P}$ where each $p_\alpha$ has coordinates $\{\theta_\alpha^1, \cdots, \theta_\alpha^m\}$. Consider the subset $C \subseteq \mathcal{P}$ such that $C$ consists of all points $q$ with coordinates $\{\theta_q^1, \cdots, \theta_q^m\}$ where

$$\theta_q^i = \sum_{\alpha=1}^n w_\alpha(q) \theta_\alpha^i, \ i = 1, \cdots, m, \text{ with}$$
$$\sum_{\alpha=1}^n w_\alpha(q) = 1, \text{ and } 0 \leq w_\alpha(q) \leq 1, \ \alpha = 1, \cdots, n. \tag{21}$$

The $n$ values $\{w_1(q), \cdots, w_n(q)\}$ are the relative weights that characterise the point $q$. Clearly any point $q \in C$ defines a probability distribution over $\mathcal{X}$ which can be given explicitly as the weighted sum

$$q(x^l) = \sum_{\alpha=1}^n w_\alpha(q) p_\alpha(x^l), \ l = 0, 1, \ldots, m. \tag{22}$$

Hence different points in $C$ are characterised by different sets of relative weights $\{w_1(q), \cdots, w_n(q)\}$ and $C$ consists of all probability distributions over $\mathcal{X}$ which can be expressed as weighted sums in the form (22). We have the following proposition:

*Proposition 1:* $C \subseteq \mathcal{P}$ is the convex hull of the set of distributions $\{p_1, \cdots, p_n\}$.

In order not to interrupt our narrative we have banished the proof to Appendix A.

In plain terms the proposition asserts that a probability distribution, which is computed as the weighted sum of a set of reference distributions, will lie in the convex hull generated by the reference set. A perusal of the proof in Appendix A would convince one that the position of the computed distribution in the convex hull is determined by the relative weights. Let us now hark back to Section III-*A*. The weighted sum algorithm (8) computes the distribution in (16) as a weighted sum of the $n$ distributions in (17). Thus the computed distribution lies in the convex hull spanned by the $n$ distributions in (17), the exact position being determined by the relative weights. Hence according to the arguments put forward in Section III-*A* the weighted sum algorithm does capture the expert's judgmental strategy.

# IV. CONCLUSION

What makes the weighted sum algorithm attractive is the fact that it is at its essence a linear process. The linear structure, as we saw in the last section, is a necessary consequence of the way we think about a set of probability distributions. Occasionally one encounters points of view expressing general concern about linear models. As far as our efforts in this paper are concerned, such qualms are unfounded as a plethora of research in the process of decision making shows. In fact research shows that linear models often outperform experts [31]. As Dawes and Corrigan write [32] - "When there are actual criterion values against which the prediction of both the judge and the linear model of the judge can be compared, the (paramorphic) linear model often does a better job than does the judge himself. That is, the correlation between the output of the model and criterion is often higher than the correlation between the decision maker's judgement and criterion, even though the model is based on the behaviour of the decision maker". The success of linear models, according to these authors, arise from the fact that, whereas linear models distil the essence of a judge's expertise they avoid the otherwise unreliable part of the judge's reasoning process, the part that is affected by extraneous factors like boredom and fatigue. In fact this is precisely the reason why a CPT produced through the weighted sum algorithm would be more consistent than one produced manually by the expert.

As indicated in the introduction, the weighted sum algorithm has been used to populate the CPTs of Bayesian networks dealing with problems related to military command and control [11]-[13]. For such problems the domain experts and the analysts are often the same. In other words an analyst would use his domain knowledge to generate the network and then use the network to make decisions. Needless to say that such analysts operate under constraints of limited time, and hence prefer a process that can be implemented within a sensible duration of time with manageable cognitive challenges. What is more the analyst needs to have confidence in the network he generates. This confidence would be lacking if the analyst were to populate the CPT, with its exponentially large number of distributions, manually. This is because of the difficulties in maintaining consistency among the distributions, across the CPT, as discussed in the introduction.

The weighted sum algorithm ameliorates this predicament in two ways. Firstly the expert is required to come up with a vastly fewer number of probability distributions as input to the algorithm. It then becomes possible to maintain consistency, and be convinced about it. Furthermore as discussed in Section II-*B*, it is often possible to exploit the symmetries in a network to drastically reduce the number of distributions required as an input to the algorithm. When such reductions are possible the problem of maintaining consistency among the subjectively produced distributions becomes almost trivial. Secondly the concept of a weighted sum is familiar to almost all decision makers. Therefore the workings of the algorithm are intuitively transparent to the analyst and thus help promote further confidence in the generated network when used for decision making.

Finally fine tuning a CPT generated through the weighted sum algorithm is rather straightforward. An analyst would do this so that the probability distributions computed, match closely up to his expectations. In most cases a slight tweak to the relative weights would suffice; in others an adjustment to the input probability distributions may be necessary. Thus a CPT can be made to evolve with the analyst's subjective experiences. The Bayesian networks we have experience with, often form parts of some larger decision making process [33]. An ideal decision making process ought to be friendly towards the process of cognition [34], and the only way a decision making software can achieve friendliness is through a process of evolution. It is this prospect of acquiring a friendly software that makes the ability to easily fine tune a CPT so appealing.

# APPENDIX A

*Proof of Proposition* 1: We prove this proposition by showing that

a). $C$ is convex and

b). $C$ is contained in every convex set that contains the $n$ distributions $\{p_1, \cdots, p_n\}$.

a): Let $a$ and $b$ be two points in $C$ having coordinates $\{\theta_a^1, \cdots, \theta_a^m\}$ and $\{\theta_b^1, \cdots, \theta_b^m\}$ respectively. From (21) it follows that these coordinates satisfy the relations:

$$\theta_a^i = \sum_{\alpha=1}^n w_\alpha(a) \theta_\alpha^i, \ i = 1, \cdots, m, \text{ with}$$
$$\sum_{\alpha=1}^n w_\alpha(a) = 1, \text{ and } 0 \leq w_\alpha(a) \leq 1, \ \alpha = 1, \cdots, n;$$
(A1)

and

$$\theta_b^i = \sum_{\alpha=1}^n w_\alpha(b) \theta_\alpha^i, \ i = 1, \cdots, m, \text{ with}$$
$$\sum_{\alpha=1}^n w_\alpha(b) = 1, \text{ and } 0 \leq w_\alpha(b) \leq 1, \ \alpha = 1, \cdots, n.$$
(A2)

Here $\{w_\alpha(a)\}$ and $\{w_\alpha(b)\}$ are the relative weights that characterise the points $a$ and $b$ respectively.

Let $\theta^i(t) : [0, 1] \to \mathcal{P}$ be the geodesic joining $a$ with $b$, i.e.,

$\theta^i(0) = \theta_a^i$ and $\theta^i(1) = \theta_b^i$, $i = 1, \cdots, m$. (A3)

To prove that $C$ is convex we have to show that the points $\{\theta^1(t), \cdots, \theta^m(t)\}$, $0 \leq t \leq 1$, all lie in $C$. Since we are working with the flat connection - all the coefficients $\Gamma_{ij,k}^{(-1)}(\theta)$ vanish - the geodesic equation reduces to [29], [30]:

$$\frac{d^2 \theta^i(t)}{dt^2} = 0$$
(A4)

Taking into account the boundary conditions (A3) the solution to (A4) takes the form

$\theta^i(t) = (1-t)\theta_a^i + t\theta_b^i$, $0 \leq t \leq 1$ and $i = 1, \cdots, m$.

Substituting from (A1) and (A2) we get

$$\theta^i(t) = (1-t)\sum_{\alpha=1}^{n} w_\alpha(a)\theta^i_\alpha + t\sum_{\alpha=1}^{n} w_\alpha(b)\theta^i_\alpha$$

$$= \sum_{\alpha=1}^{n} w_\alpha(t)\theta^i_\alpha$$

where $w_\alpha(t) = (1-t)w_\alpha(a) + tw_\alpha(b)$

From (A1) and (A2) it follows that

$$\sum_{\alpha=1}^{n} w_\alpha(t) = 1, \text{ and } 0 \leq w_\alpha(t) \leq 1, \forall \alpha.$$

Hence the points $\{\theta^1(t),\cdots,\theta^m(t)\}, 0 \leq t \leq 1$, all lie in $C$, hence $C$ is convex.

b): Let $\mathcal{D}$ be a convex set that contains the $n$ distributions $\{p_1,\cdots,p_n\}$. Since $\mathcal{D}$ contains $p_1$ and $p_2$ it contains all the points on the geodesic joining $p_1$ with coordinates $\{\theta^1_1,\cdots,\theta^m_1\}$ and $p_2$ with coordinates $\{\theta^1_2,\cdots,\theta^m_2\}$.

Let $\theta^i(t) : [0,1] \to \mathcal{P}$ be this geodesic. We have

$$\theta^i(0) = \theta^i_1 \text{ and } \theta^i(1) = \theta^i_2, \; i = 1, \cdots, m. \tag{A5}$$

Taking the boundary conditions (A5) into account, the solution to (A4) takes the form

$$\theta^i(t) = (1-t)\theta^i_1 + t\theta^i_2, \; 0 \leq t \leq 1 \text{ and } i = 1, \cdots, m.$$

Hence a typical point on the geodesic joining $p_1$ and $p_2$ has coordinates $w_1\theta^i_1 + w_2\theta^i_2$, $i = 1, \cdots, m$, for some $w_1$ and $w_2$, for which $w_1 + w_2 = 1$ and $0 \leq w_1, w_2 \leq 1$. All such points are contained in $\mathcal{D}$.

We intend to invoke induction. Suppose that for some $r < n$, $\mathcal{D}$ contains all the points of the form

$$\sum_{\alpha=1}^{r} w_\alpha \theta^i_\alpha, \; i = 1, \cdots, m, \text{ for some } 0 \leq w_1,\cdots,w_r \leq 1, \text{ for which } \sum_{\alpha=1}^{r} w_\alpha = 1 \tag{A6}$$

The above points are associated with the distributions $p_1,\cdots,p_r$, since $\mathcal{D}$ also contains the distribution $p_{r+1}$ it follows that it contains all the points on the geodesic joining any point of the type (A6) to the point $\{\theta^1_{r+1},\cdots,\theta^m_{r+1}\}$. The geodesic equation (A4) with such boundary conditions gives the solution

$$\theta^i(t) = (1-t)\sum_{\alpha=1}^{r} w_\alpha \theta^i_\alpha + t\theta^i_{r+1}, \; 0 \leq t \leq 1 \text{ and } i = 1, \cdots, m.$$

Hence a typical point on this geodesic has coordinates $\sum_{\alpha=1}^{r+1} w_\alpha \theta^i_\alpha$, $i = 1, \cdots, m$, for some weights $0 \leq w_1,\cdots w_{r+1} \leq 1$, for which $\sum_{\alpha=1}^{r+1} w_\alpha = 1$ and all such points are contained in $\mathcal{D}$. Invoking induction we conclude that $\mathcal{D}$ contains all the points of the type $\sum_{\alpha=1}^{n} w_\alpha \theta_\alpha^i$, $i = 1, \cdots, m$, for some $0 \leq w_1, \cdots w_n \leq 1$, for which $\sum_{\alpha=1}^{r} w_n = 1$. Now referring back to (21) we see that $\mathcal{D} \supseteq C$. This combined with the fact that $C$ is convex proves *Proposition* 1.

ACKNOWLEDGEMENT

The author wishes to thank Mike Davies and Lucia Falzon for many insightful discussions and John Dunn for writing the initial computer programs. Thanks are also due to Jayson Priest for his part in building the software implementing the algorithm, and obtaining valuable user feedback.